\documentclass[journal, letterpaper]{IEEEtran}
\usepackage{graphicx}
\usepackage{url}      
\usepackage{multicol}
\usepackage{subfigure}
\usepackage{amsmath}
\usepackage{amssymb}
\usepackage{cite}
\usepackage{amsfonts}
\usepackage{graphicx}
\usepackage{url}
\usepackage{ccaption}
\usepackage{booktabs} 
\usepackage{stfloats}
\usepackage{multirow}
\usepackage{amsmath}   
\usepackage{multicol}
\usepackage{multirow}
\usepackage{textgreek}
\usepackage{listings}
\usepackage{longtable}

\begin{document}

	\title{Lightweight single-image super-resolution network based on dual paths}
	\author{Ke Li, Yukai Liu}
	\maketitle

\begin{abstract}
Currently, there are two main mainstream models for single-image super-resolution (SISR) algorithms under deep learning, one is based on convolutional neural network (CNN) and the other is based on Transformer.The former designs the model by stacking convolutional layers with different convolutional kernel sizes, so that the model can better extract the local features of the image; the latter designs the model by using the self-attention mechanism, through which the self-attention The latter uses the self-attention mechanism to design a model that establishes long-distance dependencies between image pixels, so as to better extract the global features of an image. However, both methods face their own problems. Based on this, a new lightweight multi-scale feature fusion network model, TCC-TransNet (or CT-FusionNet), is proposed in this paper, which fuses the respective features of Transformer and Convolutional Neural Networks through a two-branch network architecture to achieve global and local information of global and local information. Meanwhile, considering the partial information loss that may be caused when deep neural networks train low-pixel images, this paper designs a modular connectivity method with multi-stage feature supplementation (MCM-FS), which fuses the feature maps extracted from the shallow stage of the model with those extracted from the deeper stage, in order to minimise the loss of feature image information conducive to image restoration, thus facilitating the acquisition of a higher-quality restored image ( RIQ). The experimental results show that compared with other lightweight models with the same number of parameters, the model proposed in this paper exhibits optimality in ×2, ×3 and ×4 image recovery performance

\end{abstract}

\section{Introduction}
\PARstart{T}{he} Single Image Super-Resolution(SISR) problem\cite{wang2020deep} is
a classical problem in the field of Image Super-Resolution. The goal
of this field is to reconstruct and recover an image from a given
low-resolution image to obtain a corresponding high-resolution
image. Currently, this technique has been widely used in medical
imaging\cite{shi2013cardiac}, remote sensing\cite{chen2023continuous}\cite{thornton2006sub}, video surveillance\cite{zou2011very} and other image fields\cite{matsui2017sketch}.

Research on image super-resolution has made profound developments in recent years, and deep learning-based image super-resolution algorithms are now widely used in single-image super-resolution tasks. The main advantage of the deep learning-based super-resolution algorithm is that it no longer needs to artificially process the data features in the early stage, but makes the model learn the mapping function between LR and HR in an end-to-end way, which substantially improves the performance of the relevant SISR model.
\begin{figure*}[ht] {\centering
      \centerline{\includegraphics[width=\textwidth]{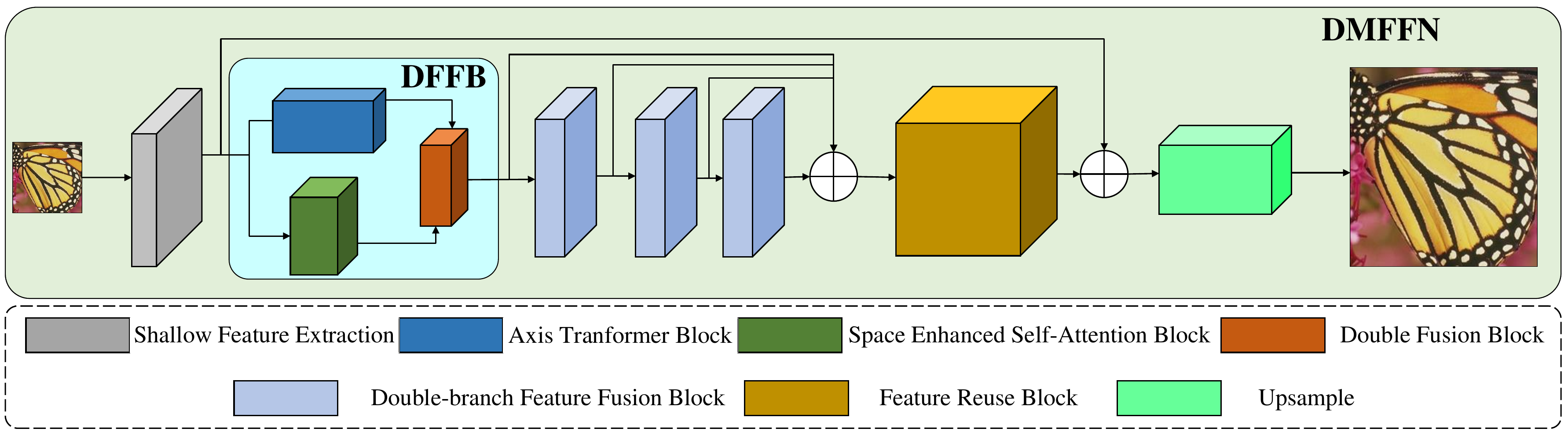}} 
      \caption{Main framework of DMFFN}
      \label{fig:enter-label}
      }
\end{figure*}
In the early days of deep learning applied to the image super-resolution problem, firstly since the work of Dong et al\cite{dong2015image}. Convolutional Neural Networks were mainly used for the image super-resolution task. However, the main problem with SISR models using convolutional neural networks is that the performance of the convolutional neural network is heavily dependent on the size of the model itself, i.e., the number of parameters and the amount of computation in the model itself. To further improve and enhance the performance of convolutional neural networks, the only way is usually to increase the network size and depth of the model. However, models with too large parameters are more demanding on the hardware devices they are deployed on, requiring strong computational and storage capabilities of the corresponding devices. Therefore models with too large network size are difficult to deploy on non-specialised devices for training. This situation is a great challenge for individuals, enterprises or research institutes that lack computational resources. Therefore, it is necessary to investigate how to lighten the image super-resolution model to reduce the number of model parameters and computation while maintaining a balance with the model performance.

In addition to this, as Transformer has achieved great success in many tasks in the field of Natural Language Processing in recent years, it has prompted people to migrate Transformer, especially techniques such as attention mechanisms, to the field of computer vision\cite{yoo2023enriched}. In this context, Liu et al\cite{liu2021swin}. proposed the Swin Transformer model, which has been widely applied to computer vision tasks due to its excellent performance. The Swin Transformer model firstly slices the image by slicing the window, then uses relative position encoding or absolute position encoding between the windows and uses the Shifting Window technique to encode the position of the window on the image. Window) technique to perform multi-level self-attention computation on the image for differently positioned windows, giving the Transformer the ability to establish long-distance pixel dependencies on the image as well. This technique enables the Transformer model to achieve performance beyond that of convolutional neural networks on computer vision tasks.

Although the Swin Transformer model has achieved great success in the field of computer vision, it still suffers from a drawback: the self-attention computation performed by the model inside the window is only local attention, and the establishment of global dependence between image pixels depends on the moving distance of the same window\cite{chen2021vision} concerning the image between different layers of the network model. Therefore, the image super-resolution network designed based on the Swin Transformer model for a small number of parameters and few network layers is unable to effectively establish a global dependency on the target image, and thus the performance still has a lot of room for improvement.

So in the design process of a lightweight single-image super-resolution network, it is difficult to use a single convolutional neural network or Transformer to achieve the performance of image recovery that people want. To solve the above problems, this paper designs a lightweight single-image super-resolution network based on dual paths, where one branch uses Transformer to extract local and regional features from the image, and one branch uses a convolutional neural network consisting of depth-separable convolutions and combines the channel attention module and spatially-enhanced attention module to extract global coarse-grained features from the image. The network interacts with the information of the features extracted by each of the two branches in the intermediate stage of the model, so that the feature maps after the information interaction retain sufficient information conducive to image recovery, thus achieving the purpose of improving the model's image recovery and image reconstruction capabilities.

On the fusion of two-way features, this paper proposes a fusion module based on multi-scale convolution. In this module by inputting the features to different convolutional branches for convolution and channel attention calculation, we can make the features of the dual-branch no longer have obvious demarcation after the fusion block processing, so that the local and global information can be fully fused. In addition, this paper combines all the intermediate stage features additively before the image reconstruction module, and it is argued that in the deep image super-resolution network, with the deepening of the depth, the network will gradually discard the low-frequency features and retain the high-frequency features, but in fact, in the image restoration, both the low-frequency features and the high-frequency features should contain a certain amount of effective information. So to make full use of the low-frequency and high-frequency information, this paper chooses to combine the features of each stage, to maximally retain the information that is effective for recovering images.

Finally, in the model image reconstruction stage, this paper uses sub-pixel convolution to upsample the features which are widely used in this field.

The final results show that the performance of the model proposed in this paper is optimal for image recovery in comparison with other lightweight models.

In summary, the main work of this paper is:

(1) An innovative two-way fusion network of Convolution and Transformer is designed, which makes full use of the characteristics of the Convolution and Transformer branches using a two-way structure, effectively improving the limitations and shortcomings of previous methods in extracting global features of the image, and significantly enhancing the performance of the lightweight super-resolution reconstruction model;

(2) An efficient convolutional network branch is constructed, which enables the branch to focus on extracting global coarse-grained information of the image and broader image context information through the orderly combination of deeply separable convolution and spatial self-attention computational modules;

(3) A novel multi-scale dual-path feature fusion block is constructed, which enables the dual-branch features to complement each other's information by splitting the input features multiple times and inputting them into multiple branches for convolution or channel attention computation followed by fusion, effectively integrating the image feature information from different architectural branches to achieve more detailed and accurate super-resolution reconstruction results;

(4) A new feature multiplexing module is designed, which retains the low and high-frequency information favourable to image recovery through the fusion of multi-stage features and further compensates for the high-frequency details that may have been lost in the above processing, thus enhancing the quality of image recovery.
 \begin{figure}
    \centering
    \includegraphics[width=1\linewidth]{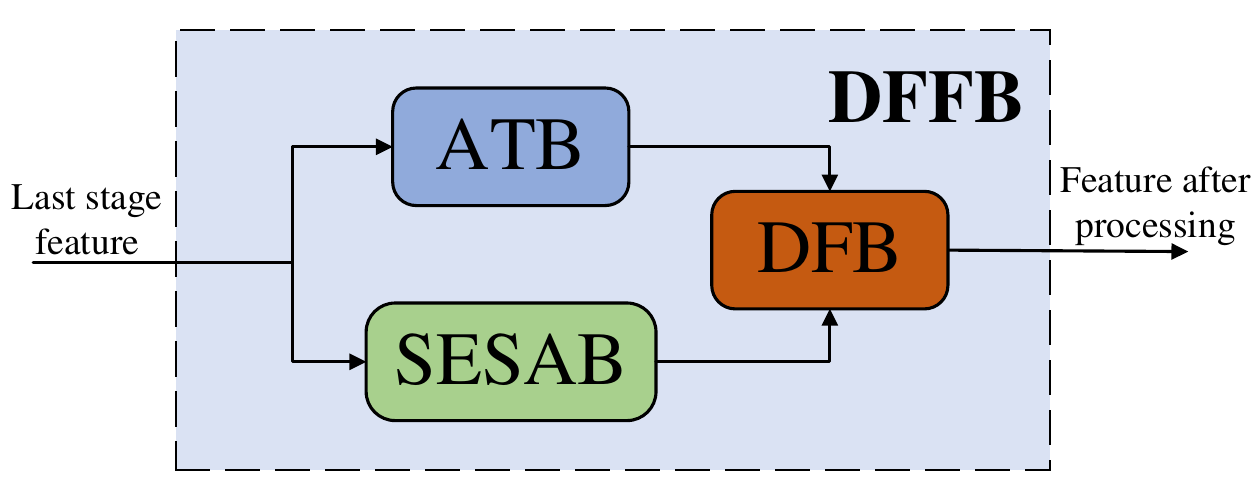}
    \caption{Mechanisms for DFFB processing features}
    \label{fig:enter-label}
\end{figure}
\begin{figure*}[ht] {\centering
      \centerline{\includegraphics[width=\textwidth]{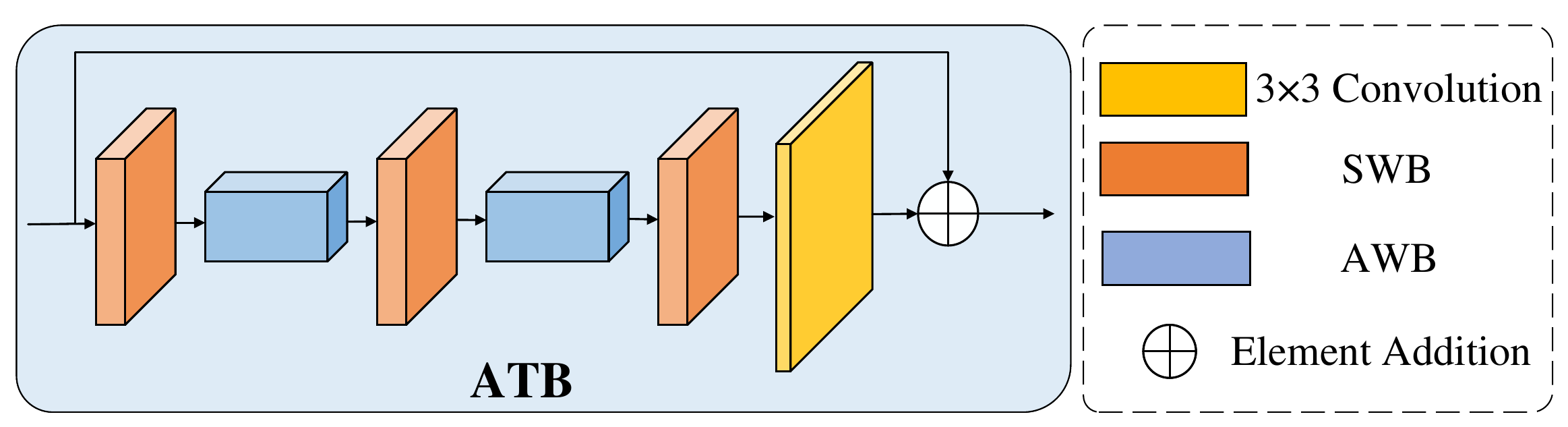}} 
      \caption{ATB network architecture}
      \label{fig:enter-label}
      }
\end{figure*}

\section{Method}
In this paper, a new dual-way lightweight image super-resolution network based on the fusion of Transformer and convolutional neural network is proposed, and the main structure is shown in Fig. 2.

The network mainly contains the following modules: shallow feature extraction block (SFB: Shallow Feature Block), dual-way feature fusion block (DFFB: Double-way Feature Fusion Block), feature reusing block (FRB: Feature Reusing Block), and double-three times upsampling block ( Upsample).

The model first makes the low-resolution image pass through a shallow feature extraction block, which consists of a $3\times3$ convolution and an ELU activation function, aiming at first extracting the shallow feature map of the image at the very beginning stage, which is convenient for subsequent feature mining. Then the model passes the feature map through some two-way feature fusion blocks for feature extraction, and finally, several intermediate features obtained through the various stages of the DFFB in front of the final additive connection are merged and passed through the feature reuse module, to facilitate the full fusion of each intermediate feature to disrupt the information, and to carry out the subsequent information supplementation. Finally, the fully fused features are passed through the upsampling module for image reconstruction.

Assuming that the input low-resolution image is denoted as $I_{LR}$, shallow feature extraction is first performed on $I_{LR}$:
\begin{equation}
    \begin{split}
        f_{0}=Conv_{3 \times 3}{(I_{LR})}
    \end{split}
\end{equation}
After that $f_{0}$ is fed into the feature extraction fusion block for feature extraction to get intermediate features for each stage:
\begin{equation}
    \begin{split}
        f_{n}=DFFB(f_{n-1}),n=0,1,2,3
    \end{split}
\end{equation}
Then each intermediate feature is spliced and input to the multiplexed feature fusion block to get the fused feature image:
\begin{equation}
    \begin{split}
        f_{4}=FRB(f_{0}+f_{1}+f_{2}+f_{3})
    \end{split}
\end{equation}
Finally, the fused image is then spliced with the shallow features and finally upsampled to get the reconstructed high-resolution image:
\begin{equation}
    \begin{split}
        I_{HR} = Upsample(f_{0}+f_{4})
    \end{split}
\end{equation}

Next, this article will talk about the internal composition and operation mechanism of each block in detail to facilitate the understanding of the operation process of the network.
\begin{figure*}[ht] {\centering
      \centerline{\includegraphics[width=\textwidth]{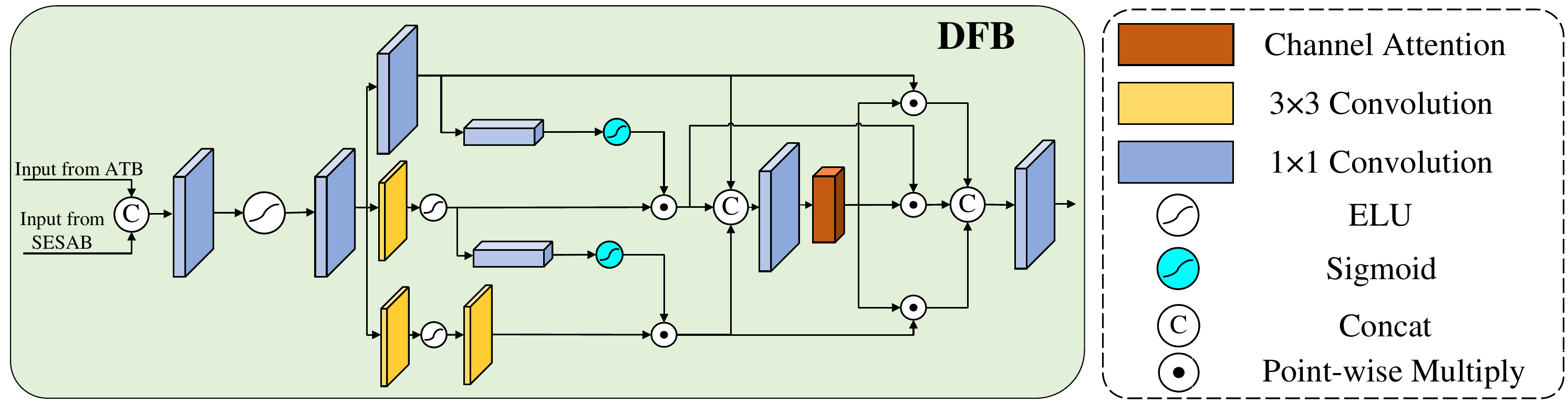}} 
      \caption{DFB network architecture}
      \label{fig:enter-label}
      }
\end{figure*}
\subsection{Double-Branch Feature Fusion Block Design}

Firstly, there is the Dual-Feature Fusion Block (DFFB), whose main structure is shown in Fig. 3.

The DFFB module consists of three functional blocks, which are Axis Transformer Block (ATB: Axis Transformer Block), Space Enhanced Self-Attention Block (SESAB: Space Enhanced Self-Attention Block) and Double-Way Fusion Block (DFB: Double-Way Fusion Block).

Firstly, there is the Axis Transformer Block (ATB), whose main structure is shown in Fig. 4. In DFFB, the main role of the ATB module is to extract the local and regional features of the image. Through the self-attention mechanism, it establishes the regional dependency between image pixel points to better extract the local low-frequency information of the image.

Then there is the Space Enhanced Self-Attention Block (SESAB: Space Enhanced Self-Attention Block), whose main structure is shown in Fig. 5.
The main role of this block in DFFB is to extract the global coarse-grained information of the image. This module helps the model to adaptively extract the critical information in the image and simultaneously weaken the non-critical information in the image during the training process through the stacking of depth-separable convolutional layers in a specific order and the combination of channel attention and spatially enhanced attention modules. This combination helps the model to efficiently extract the global coarse-grained information, which in turn compensates for the information in the branch where the ATB module is located.

Finally, there is the Double-Way Fusion Block (DFB: Double-Way Fusion Block), the main structure of which is shown in Fig. 6. From Fig. 3, we can know that after the feature map of the previous stage enters the DFB, it enters the ATB and SESAB respectively to be processed, and finally all the local area and global features extracted from the two branches are put into the DFB to be fused, and the information of its two branches is fully disrupted and mixed, and finally inputted into the next stage to be processed. Its formula can be expressed as follows:
\begin{equation}
    \begin{split}
        f_{i} = DFB(ATB(f_{i-1}),SESAB(f_{i-1}))
    \end{split}
\end{equation}
Equation 6 indicates that the feature $f_{i}$ of the $i$th stage is obtained from the feature $f_{i-1}$ outputted from the previous DFB module after processing by ATB, SESAB module and then fused by DFB module.

\subsubsection{Axis Transformer Block}
The Swin Transformer model proposed by Liu et al\cite{liu2021swin}. has achieved great success in the field of computer vision, which achieves the effect of establishing long-distance dependence on the more distant pixels of an image by slicing the image, partitioning it into image blocks (Patch) with partial overlap, and then using a multi-layered square window to compute the self-attention between the blocks to achieve the effect of establishing long-distance dependence on the more distant pixels of the image. However, the model requires a larger number of network layers to slice the image to achieve the effect of square windows moving across the graph, and thus the model requires a larger amount of computation to take advantage of its global modelling capabilities. Some previous works tried to reduce the number of layers of the Swin Transformer for feature extraction, but the experimental results found that once the number of layers of the Swin Transformer is reduced, the global feature extraction ability of the model will be greatly reduced, and it will be more biased towards extracting regional and local features of the image.

In order to solve the above problems, a new self-attention computational mechanism is used in this paper, as shown in Fig. 7.

This self-attention computational mechanism changes the shape of the self-attention window and uses an axial attention window (the red part) to compute the self-attention. The advantage of using an axial window is that the window can span multiple square windows, increasing the number of indirectly dependent pixels to be established while decreasing the number of pixels for which direct dependencies are computed, thus allowing the model to quickly establish global dependencies of pixels with fewer network layers.

Based on the axial self-attention computational mechanism of Fig. 7, this algorithm uses the Axial Transformer computational module to extract the image's regional and local features as shown in Fig. 4.
Where AWB(Axis Window Block) module is the axial attention computing module and its internal structure is shown in Fig. 8.
\begin{figure}
    \centering
    \includegraphics[width=1\linewidth]{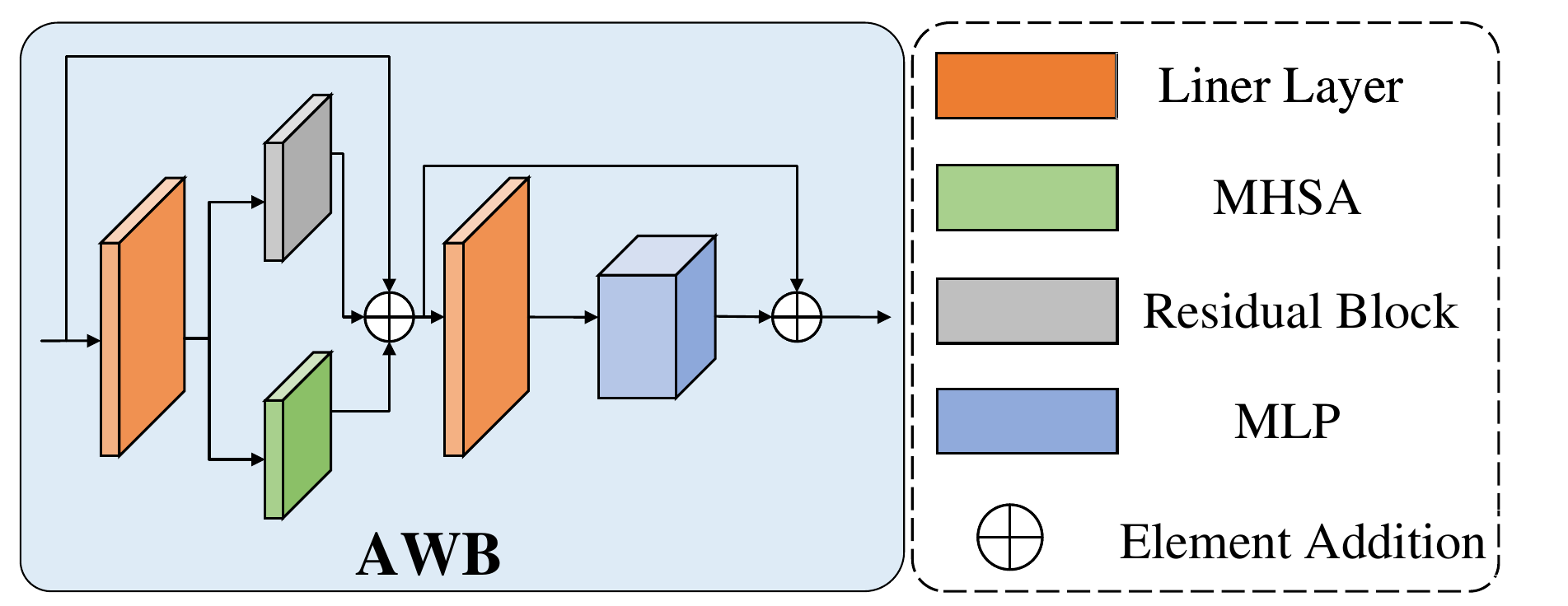}
    \caption{AWB Block}
    \label{fig:enter-label}
\end{figure}
SWB(Square Window Block) module is the square window self-attention computing module with the same mechanism as Swin Transformer, and its internal structure is shown in Fig. 9.
\begin{figure}
    \centering
    \includegraphics[width=1\linewidth]{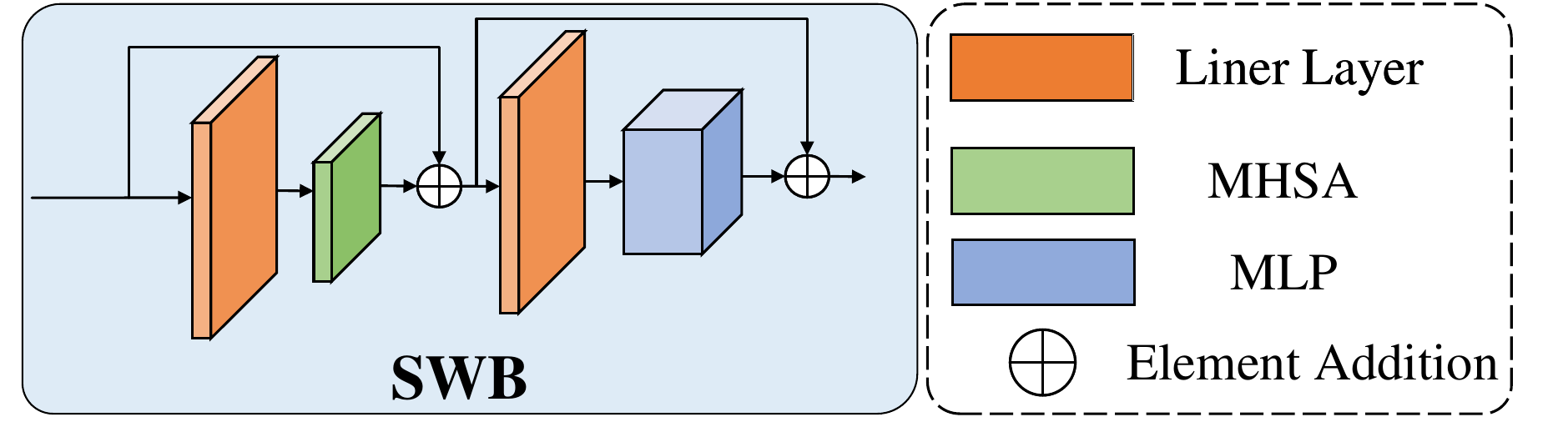}
    \caption{SWB Block}
    \label{fig:enter-label}
\end{figure}
Both the AWB module and SWB module are self-attention computation modules constructed using linear layers, fully connected layers, Multi-Head Self-Attention (MHSA: Multi-Head Self-Attention), and Residual Block. For the features $f_{i}$ entering the AWB.
\begin{equation}
    \begin{split}
        f_{i+1}=f_{i}+RB(LN(f_{i}))+MHSA(LN(f_{i}))
    \end{split}
\end{equation}
\begin{equation}
    \begin{split}
        f_{i+2}=f_{i+1}+MLP(LN(f_{i+1}))
    \end{split}
\end{equation}
Finally, $f_{i+2}$ is obtained as the output feature.

For the features $f_{i}$ entering the SWB.
\begin{equation}
    \begin{split}
        f_{i+1}=f_{i}+MHSA(LN(f_{i}))
    \end{split}
\end{equation}
\begin{equation}
    \begin{split}
       f_{i+2}=f_{i+1}+MLP(LN(f_{i+1}))
    \end{split}
\end{equation}
Finally, get $f_{i+2}$ as the output feature.
\begin{figure*}[ht] {\centering
      \centerline{\includegraphics[width=\textwidth]{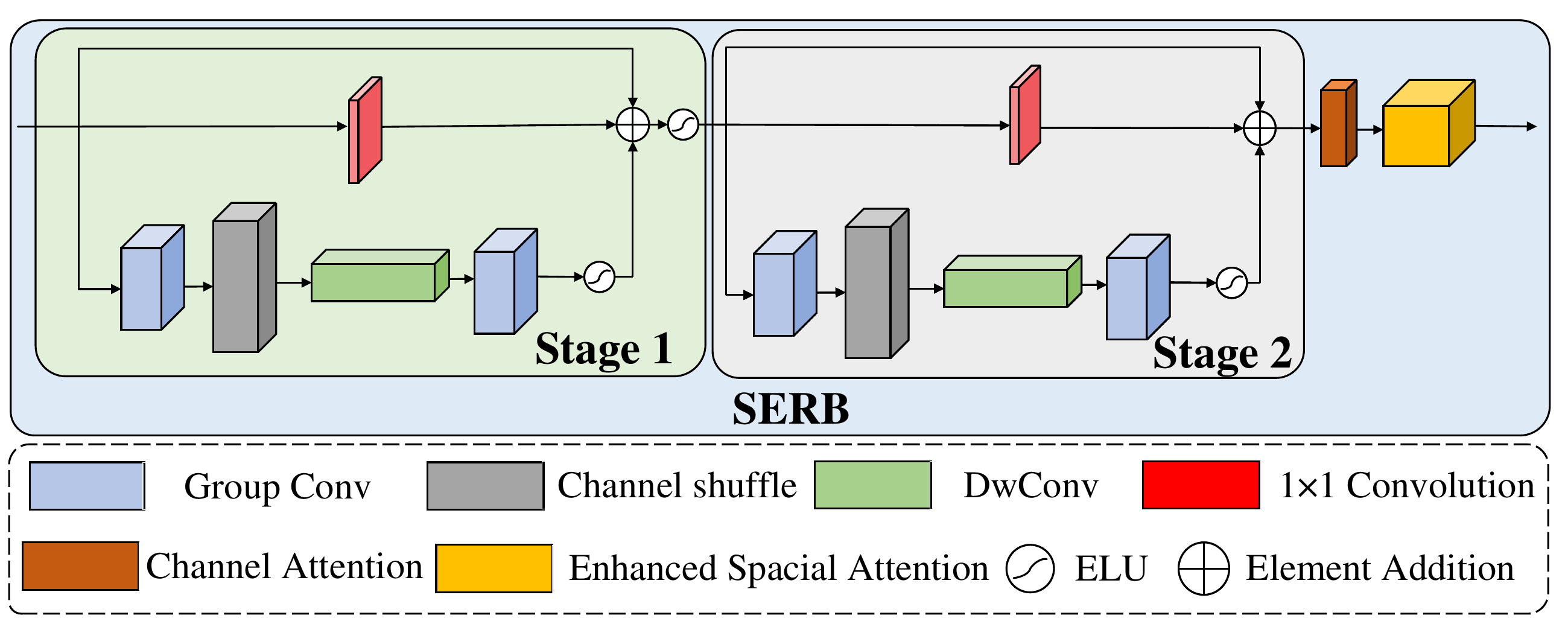}} 
      \caption{SERB main framework}
      \label{fig:enter-label}
      }
\end{figure*}
The method replaces the moving window mechanism of the Swin Transformer by designing the cross-computation of the axial attention window and square window, which drastically reduces the computation of the Transformer branch and efficiently establishes the long-distance indirect dependency between pixels.
\begin{figure*}[ht] {\centering
      \centerline{\includegraphics[width=\textwidth]{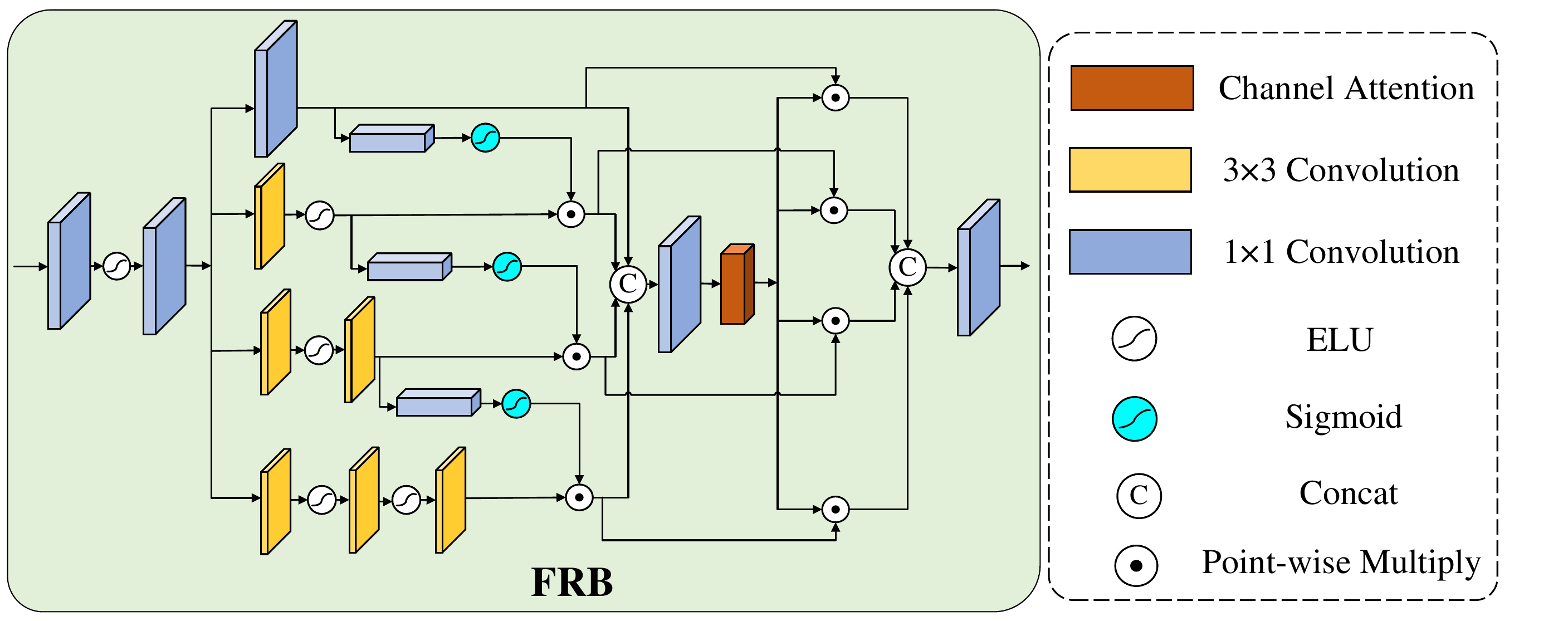}} 
      \caption{FRB main framework}
      \label{fig:enter-label}
      }
\end{figure*} 

\subsubsection{Space Enhanced Self-Attention Block}
In the design of the convolutional branch, this model refers to the construction of RCAN\cite{lin2022revisiting}, by using the designed residual convolution block (SERB: Space Enhanced Residual Block) stacking to achieve the convolutional neural network on the image feature extraction. At the same time, in order to alleviate the rising number of model parameters due to the stacking of residual convolutional blocks, the part uses depth-separable convolution to do a lot of lightweight processing of residual convolutional blocks, and innovatively combines the spatial enhanced attention module (ESA: Enhanced Self-Attention)\cite{liu2020residual} and channel attention module (CA: Channel Attention)\cite{hu2018squeeze} at the end of each residual convolutional block, which is used to extract features from images by the convolutional neural network. Channel Attention), which is used to enhance the ability of the residual convolution block to extract image features, so that the residual convolution block can have the ability to extract the global coarse-grained features of the image, which is convenient for the subsequent supplementation of the feature maps processed by the ATB module.

The main structure of SERB is shown in Fig. 10:

Suppose the input features are $f_{i}$:
\begin{equation}
    \begin{split}
        f_{i3}=ELU(GConv(DwConv(GConv(f_{i}))))
    \end{split}
\end{equation}
\begin{equation}
    \begin{split}
        f_{i+1}=ELU(f_{i}+Conv(f_{i})+f_{i3})
    \end{split}
\end{equation}
where $f_{i3}$ denotes the feature map of the output of the third branch counting down from the top in SERB stage 1.

For stage 2, this paper uses $f_{(i+1)3}$ to denote the feature map of the output of the third branch counting down from the top in SERB stage 2, and so there is:
\begin{small}
    \begin{equation}
    \begin{split}
        f_{(i+1)3}=ELU(GConv(DwConv(GConv(f_{i+1}))))
    \end{split}
\end{equation}
\end{small}
\begin{small}
    \begin{equation}
    \begin{split}
        f_{i+2}=ESA(CA((f_{i+1}+Conv(f_{i+1})+f_{(i+1)3}))))
    \end{split}
\end{equation}
\end{small}
Thus the output feature of the SERB is $f_{i+2}$.

In the construction process of the above SERB, the model uses the spatially enhanced attention module and the channel attention module. The main role of the channel attention module is to give each channel of the feature map a separate adaptive feature weight during the training process so that it can achieve the effect of weakening the invalid features and strengthening the important features by adaptively adjusting the feature weights during the training process, and improve the efficiency and quality of the model's image recovery. After processing by the channel attention module, the spatial enhancement attention module can further extract the global coarse-grained features of the image, thereby improving the effect of model pixel recovery.

\subsection{Feature Reuse Block}
When the depth of the network is deeper, it was examined during the experiments that the feature image loses some of the information that helps in image recovery as the training progresses in the deep network. Some previous studies have also shown that both shallow feature images and deep feature images have the same information that helps image recovery for reconstruction.

In view of the phenomenon that the quality of the reconstructed image is degraded due to the loss of information in the shallow layer of the feature image, a new feature reuse module is specially designed in this model before the image reconstruction stage. And this technique adopts the same mechanism as DFB, using four branches of progressive feature fusion to fully fuse the shallow features with the deeper features, fully mixing the useful information in the shallow feature map and the information in the deeper feature map that is beneficial to the image recovery, to achieve the maximisation of the efficiency of the image reconstruction and the minimisation of the loss of the feature information. The internal structure of the FRB module is shown in Fig. 11.

The model adds the intermediate feature images of the previous four stages in the learning process and then inputs them into the FRB module for the deep fusion and restructuring of the shallow features and the deep features, to avoid the loss of information in the training process as much as possible.

\begin{table*}[ht]
  \centering
  \caption{Quantitative comparison (average PSNR/SSIM) with state-of-the-art methods for \textbf{lightweight image SR} on benchmark datasets.}
  \resizebox{\linewidth}{!}{
    \begin{tabular}{|c|c|c|c|c|c|c|c|c|c|c|c|c|c|}
    \hline
    \multirow{2}{*}{scale} & \multicolumn{1}{c|}{\multirow{2}{*}{model}} & \multicolumn{1}{c|}{\multirow{2}{*}{param(/K)}} & \multirow{2}{*}{training set} & \multicolumn{2}{c|}{Set5} & \multicolumn{2}{c|}{Set14} & \multicolumn{2}{c|}{BSD100} & \multicolumn{2}{c|}{Urban100} & \multicolumn{2}{c|}{Manga109} \\
\cline{5-14}          & \multicolumn{1}{c|}{} & \multicolumn{1}{c|}{} &       & PSNR  & SSIM  & PSNR  & SSIM  & PSNR  & SSIM  & PSNR  & SSIM  & PSNR  & SSIM \\
    \hline
    \multirow{7}{*}{$\times$ 2} & LatticeNet\cite{rosu2022latticenet} & 800K  & DIV2K & 38.1  & 0.9608 & 33.6  & 0.9177 & 32.21 & 0.9001 & 32.29 & 0.9297 & 38.81 & 0.9773 \\
          & AWSRN-M\cite{wang2019lightweight} & 756K  & DIV2K & 38.06 & 0.9607 & 33.7  & 0.9187 & 32.2  & 0.8999 & 32.25 & 0.9288 & 38.94 & 0.9774 \\
          & MAFFSRN-L\cite{muqeet2020multi} & 1063K & DIV2K & 38.04 & 0.9605 & 33.66 & 0.9181 & 32.21 & 0.9   & 32.23 & 0.9294 & 38.66 & 0.9772 \\
          & SwinIR\cite{liang2021swinir} & 878K  & DIV2K & 38.14 & 0.9611 & 33.86 & 0.9206 & \textbf{32.31} & \textbf{0.9012} & 32.76 & 0.934 & 39.12 & \textbf{0.9783} \\
          & ELAN\cite{zhang2022efficient}  & 582K  & DIV2K & 38.17 & 0.9611 & \textbf{33.94} & \textbf{0.9207} & 32.3  & \textbf{0.9012} & 32.76 & 0.934 & 39.11 & 0.9782 \\
          & DLGSANet-tiny\cite{li2023dlgsanet} & 566K  & DIV2K & 38.16 & 0.9611 & 33.92 & 0.9202 & 32.26 & 0.9007 & \textbf{32.82} & \textbf{0.9343} & 39.14 & 0.9777 \\
          & DMFFN (ours) & 669K  & DIV2K & \textbf{38.2} & \textbf{0.9613} & 33.87 & 0.9199 & 32.29 & 0.9011 & 32.76 & 0.9336 & \textbf{39.21} & 0.9782 \\
    \hline
    \multirow{8}{*}{$\times$ 3} & LatticeNet & 765K  & DIV2K & 34.4  & 0.9272 & 30.32 & 0.8416 & 29.1  & 0.8049 & 28.19 & 0.8513 & 33.63 & 0.9442 \\
          & AWSRN-M & 1143K & DIV2K & 34.42 & 0.9275 & 30.32 & 0.8419 & 29.13 & 0.8059 & 28.26 & 0.8545 & 33.64 & 0.945 \\
          & MAFFSRN-L & 807K  & DIV2K & 34.45 & 0.9277 & 30.4  & 0.8432 & 29.13 & 0.8061 & 28.26 & 0.8552 &\multicolumn{1}{c|}{-/-} & \multicolumn{1}{c|}{-/-}\\
          & EMASRN & 437K  & DIV2K & 34.36 & 0.9264 & 30.3  & 0.8411 & 29.05 & 0.8035 & 28.04 & 0.8493 & 33.43 & 0.9433 \\
          & SwinIR & 886K  & DIV2K & 34.62 & 0.9289 & 30.54 & 0.8463 & 29.2  & 0.8082 & 28.66 & 0.8624 & 33.98 & 0.9478 \\
          & ELAN  & 590K  & DIV2K & 34.61 & 0.9288 & 30.55 & 0.8463 & 29.21 & 0.8081 & 28.69 & 0.8624 & 34    & 0.9478 \\
          & DLGSANet-tiny & 572K  & DIV2K & \textbf{34.63} & 0.9288 & \textbf{30.57} & 0.8459 & 29.21 & 0.8083 & 28.69 & 0.863 & 34.1  & 0.948 \\
          & DMFFN (ours) & 708K  & DIV2K & 34.62 & \textbf{0.9293} & 30.48 & \textbf{0.8467} & \textbf{29.26} & \textbf{0.8094} & \textbf{28.78} & \textbf{0.8645} & \textbf{34.31} & \textbf{0.949} \\
    \hline
    \multirow{8}{*}{$\times$ 4} & LatticeNet & 777K  & DIV2K & 32.18 & 0.8943 & 28.61 & 0.7812 & 27.57 & 0.7355 & 26.14 & 0.7844 & 30.54 & 0.9075 \\
          & AWSRN-M & 1254K & DIV2K & 32.21 & 0.8954 & 28.65 & 0.7832 & 27.6  & 0.7368 & 26.15 & 0.7884 & 30.56 & 0.9093 \\
          & MAFFSRN-L & 830K  & DIV2K & 32.2  & 0.8953 & 28.62 & 0.7822 & 27.59 & 0.737 & 26.16 & 0.7887 & 30.33 & 0.9069 \\
          & EMASRN & 558K  & DIV2K & 32.17 & 0.8948 & 28.57 & 0.7809 & 27.55 & 0.7351 & 26.01 & 0.7838 & 30.41 & 0.9076 \\
          & SwinIR & 897K  & DIV2K & 32.44 & 0.8976 & 28.77 & 0.7858 & 27.69 & 0.7406 & 26.47 & 0.798 & 30.92 & 0.9151 \\
          & ELAN  & 601K  & DIV2K & 32.43 & 0.8975 & 28.78 & 0.7858 & 27.69 & 0.7406 & 26.54 & 0.7982 & 30.92 & 0.915 \\
          & DLGSANet-tiny & 581K  & DIV2K & \textbf{32.46} & \textbf{0.8993} & \textbf{28.79} & 0.7871 & 27.7  & 0.7415 & 26.55 & \textbf{0.8033} & 30.98 & 0.9161 \\
          & DMFFN (ours) & 765K  & DIV2K & 32.41 & 0.8981 & 28.74 & \textbf{0.7874} & \textbf{27.74} & \textbf{0.7421} & \textbf{26.67} & 0.8029 & \textbf{31.24} & \textbf{0.917} \\
    \hline
    \end{tabular}
    }
  \label{tab:addlabel}
\end{table*}

\subsection{Results on Image SR}
\subsubsection{Quantitative comparison}
In the objective evaluation index comparison, the experimental results were selected to compare the model test results of average PSNR and average SSIM with the magnification of $\times2$, $\times3$, and $\times4$, as shown in Table 1. This comparison compares the present model with the classical models in the last three years. From the results, it can be seen that the tested values of average PSNR and average SSIM of the present model exceed the majority of the classical models and maintain a more excellent level. From the result, it can be concluded that the network architecture of the present model reaches the level of SOTA (State-Of-The-Art).

\subsubsection{Visual Comparison}
The images in the Set14 test set and the images in the Urban100 test set were selected as samples for the experimental results, as shown in Fig. 12. The original HR image, the model CARN\cite{ahn2018fast}, IMDN\cite{hui2019lightweight}, AWSRN-S\cite{wang2019lightweight}, LAPAR-A\cite{li2020lapar}, EDSR\cite{lim2017enhanced}, and SwinIR\cite{liang2021swinir} were used as the control models to show the subjective effect of the model for image recovery. From the comparison, it can be seen that the model shows significant improvement in its effectiveness for image detail texture recovery. In the Barbara image, the model can easily distinguish the texture details in the diagonal upward direction, and can also clearly distinguish the distance for the thinly spaced stripes. In the img085 image, although the reconstructed image of this model still has some distance and blurring compared with the original HR image, compared with the restoration effect of other models, the model in this paper has richer and more realistic details in the restoration of the distant light, and there is no problem of missing details, while the other images have varying degrees of missing details.

\section{Conclusion}
In this paper, a lightweight progressive multi-scale feature fusion network based on a two-brach convolutional neural network and Transformer is proposed. The main work of this paper is to design a two-way progressive feature multiscale fusion block that combines the spatially enhanced attention block to the traditional single-branch convolutional network and to fuse the local features of the image extracted by the Transformer branch with the coarse-grained global features extracted by the convolutional branch, to achieve the final information utilisation rate and the expressive ability of the model. The experimental results show that the method of this paper is effective, whether from the principle of algorithm design or specific objective experiments, it proves that the model is significantly better than the existing methods.
\bibliographystyle{unsrt}
\bibliography{ref}
\end{document}